\documentclass{article}

\usepackage{microtype}
\usepackage{graphicx}
\usepackage{subfigure}
\usepackage{booktabs} % for professional tables
\usepackage{textcomp}

\usepackage{hyperref}

\usepackage[accepted]{icml2024}

\usepackage{amsmath}
\usepackage{amssymb}
\usepackage{mathtools}
\usepackage{amsthm}

\usepackage[capitalize,noabbrev]{cleveref}

\theoremstyle{plain}

\theoremstyle{definition}

\theoremstyle{remark}

\usepackage[textsize=tiny]{todonotes}

\icmltitlerunning{Group Crosscoders}

\begin{document}

\twocolumn[
\icmltitle{Group Crosscoders for Mechanistic Analysis of Symmetry}



\icmlsetsymbol{equal}{*}

\begin{icmlauthorlist}
\icmlauthor{Liv Gorton}{yyy}
\end{icmlauthorlist}

\icmlaffiliation{yyy}{Independent}

\icmlcorrespondingauthor{Liv Gorton}{liv@livgorton.com}

\icmlkeywords{mechanistic interpretability, InceptionV1, sparse autoencoders, crosscoders, superposition, vision interpretability, group theory}

\vskip 0.3in
]



\printAffiliationsAndNotice{} % otherwise use the standard text.

\begin{abstract}
We introduce group crosscoders, an extension of crosscoders that systematically discover and analyse symmetrical features in neural networks. While neural networks often develop equivariant representations without explicit architectural constraints, understanding these emergent symmetries has traditionally relied on manual analysis. Group crosscoders automate this process by performing dictionary learning across transformed versions of inputs under a symmetry group. Applied to InceptionV1's mixed3b layer using the dihedral group $\mathrm{D}_{32}$, our method reveals several key insights: First, it naturally clusters features into interpretable families that correspond to previously hypothesised feature types, providing more precise separation than standard sparse autoencoders. Second, our transform block analysis enables the automatic characterisation of feature symmetries, revealing how different geometric features (such as curves versus lines) exhibit distinct patterns of invariance and equivariance. These results demonstrate that group crosscoders can provide systematic insights into how neural networks represent symmetry, offering a promising new tool for mechanistic interpretability.

\end{abstract}

\section{Introduction}

Given the symmetries that exist in the natural world, it's natural to think of parallel symmetries in neural network representation. Modern deep learning's initial success was based on incorporating translational symmetry into neural network architectures in the form of convolutional neural networks (CNNs) \cite{lecun1995convolutional}. In particular, CNNs incorporate \emph{equivariance}, meaning that translation of the input corresponds to a simple transformation of the activations (in this case, also translation). Naturally, the success of CNNs led to an interest in extending neural network architectures to other symmetries \cite{bergstra2011statistical,dieleman2016exploiting}. This culminated in group convolutional neural networks, which support arbitrary group symmetries \cite{cohen2016group}.

But most neural network architectures don't explicitly incorporate these other symmetries. Even when convolutional neural networks are used, it's rare to explicitly enforce other symmetries. This leads to a natural question: to what extent do neural networks represent symmetry when it isn't enforced? \citet{lenc2015understanding} found evidence of equivariance by studying linear maps between representations of transformed inputs. But this still leaves a question: what precisely is being represented equivariantly? \citet{cammarata2020curve} demonstrate one example of equivariance at the level of individual neurons: a set of neurons which all detect curves but in different orientations. As the stimuli rotate, each neuron would ``hand off'' its activation to another curve detector neuron. Likewise, \citet{schubert2021highlow} find ``high-low frequency detector'' neurons, which also detect the same patterns in different orientations. \citet{olah2020naturally} documents a variety of such equivariant neuron families, with corresponding equivariant circuits.

These kinds of mechanistic analyses of neural networks show that equivariant structure - of the exact kind one might have thought to enforce - naturally forms in neural networks, even when it isn't enforced. However, mechanistic interpretability has progressed a long way since these analyses, and analysing individual neurons is no longer favoured. In particular, it is now widely believed that many neural network features are in \emph{superposition} \cite{arora2018linear,elhage2022superposition}, using combinations of neurons to allow the model to express more features than it has individual neurons. Over the last two years, there's been significant success using methods based on dictionary learning, and in particular sparse autoencoders, to extract features from superposition \cite{bricken2023monosemanticity,cunningham2023sparse}. \citet{gorton2024missingcurvedetectorsinceptionv1} returned to the question of the rotationally-equivariant curve detectors with this modern machinery and found many additional curve detectors. (In fact, there's evidence that such curve detector features form a manifold \cite{gorton2024curvemanifolds}.)

Crosscoders \cite{lindsey2024crosscoders} are a new extension to sparse autoencoders. Originally introduced to find analogous features across layers, or even across different models, they can also be used to find the \emph{same features across transformations}. Rather than discovering such symmetries post hoc, we can systematise them into the dictionary learning process.

We call this method a \emph{group crosscoder}. It turns out that this has many advantages. We can automatically detect equivariant features and the type of symmetry or invariances they have and automatically group them with their transformed counterparts. Additionally, we can separate aspects of feature geometry that correspond to equivariance from those that do not. This allows us to begin to systemise the notion of a feature family, automatically discovering many of the families found by \citet{olah2020an}.

\section{Methods}
\begin{figure*}[hbt!]
\begin{center}
\centerline{\includegraphics[width=\textwidth]{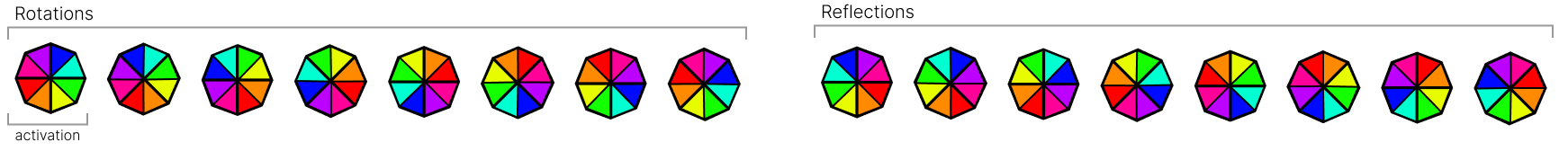}}
\caption{A representation of how a dataset example for $\mathrm{D}_8$ would be structured. Each octagon represents one set of activations from the source model. Rotations rotate in a counter-clockwise direction, whereas the reflections rotate in a clockwise direction.}
\label{figure:vector_layout}
\end{center}
\end{figure*}

Group crosscoders are an extension of crosscoders \cite{lindsey2024crosscoders}. Where crosscoders are trained on activations from different layers, or from different models, group crosscoders are trained on activations from transformed versions of the same input. 

For example, a crosscoder applied to the cross-layer use case might perform dictionary learning on vectors:

$$x = [a^l(I) ~:~ l < L]$$
where $a^l(I)$ represents the activations of the model at layer $l$ in response to image $I$. In contrast, a group crosscoder operates on vectors of the form:

$$x = [a^l(gI) ~:~ g\in G]$$

where $G$ is a group with some action on $I$. Put another way, it is the vector of activations in response to the orbit $\mathcal{G}(\mathbf{I})$ of $I$ under $G$.  In the case of this paper, we'll focus on the case where $G$ is the dihedral group of order 32 (i.e., 16 rotations), with actions rotating and flipping the image. One subtlety is that we'll take activations at a single point in the image (rather than the full convolutional activations) and track the analogous rotated and flipped point. We'll also have the crosscoder predict this vector only from the the activations in response to the untransformed image.

Note that these concatenated activation vectors can be thought of as blocks of activations, parametrized by group elements, as seen in Figure \ref{figure:vector_layout}.

\subsection{Architecture}

We now provided a more formal definition of a group crosscoder. Let $\mathrm{G}$ be the group of transformations, $n = |\mathbf{a}^l(\mathbf{I})|$ be the size of an activation for a single transformation, and $|\mathrm{G}|$ be the number of transformations in the group. Let

$$x_0 = [a^l(I) ]$$
$$x = [a^l(gI) ~:~ g\in G]$$

Then a $G$-crosscoder with $m$ features is defined:
% $$
% \begin{flalign}
% &\mathbf{f}(\mathbf{x}_0) = \mathrm{ReLU}(W_e\mathbf{x}_0 + \mathbf{b}_e) \\ &\hat{\mathbf{x}} = W_d\mathbf{f}(\mathbf{x}_0) + \mathbf{b}_d \\
% &\mathcal{L} = \frac{1}{|X|}\sum_{\mathbf{x} \in X}||\mathbf{x} - \hat{\mathbf{x}}||_2^2 + \lambda \sum_i |\mathbf{f}_i(\mathbf{x}_0)|||W_{d,i}||_2
% \end{flalign}
% $$
\begin{align*}
& \mathbf{f}(\mathbf{x}_0) = \mathrm{ReLU}(W_e\mathbf{x}_0 + \mathbf{b}_e) \\ 
& \hat{\mathbf{x}} = W_d\mathbf{f}(\mathbf{x}_0) + \mathbf{b}_d \\ 
& \mathcal{L} = \frac{1}{|X|}\sum_{\mathbf{x} \in X}||\mathbf{x} - \hat{\mathbf{x}}||_2^2 + \lambda \sum_i |\mathbf{f}_i(\mathbf{x}_0)|||W_{d,i}||_2
\end{align*}
where $W_e \in \mathbb{R}^{m \times n}$ are the encoder weights, $W_d \in \mathbb{R}^{|G|n \times m}$ are the decoder weights, $\mathbf{b}_e \in \mathbb{R}^m$ is the encoder bias, and $\mathbf{b}_d \in \mathbb{R}^{|G|n}$ is the decoder bias.

\subsection{Dataset \& Training}\label{section:dataset-training}

The dataset, $X$, for $\mathrm{G}$-crosscoders was constructed over ImageNet \cite{deng2009imagenet, imagenet15russakovsky}. Activations were collected from the fifth layer (\texttt{mixed3b}) of the convolutional neural network, InceptionV1 \cite{DBLP:journals/corr/SzegedyLJSRAEVR14}. InceptionV1 was selected due to its early layers being well-characterised in previous work, providing insight into what group structures might link the features of these layers. 

We sample $n$ activations for each image, chosen based on the untransformed image. We define a sampling function $\mathcal{S}_n(\cdot)$ that selects $n$ activations without replacement, where the probability of being chosen is proportional to the L2 norm. For each transformation of that same image, we preserve the coordinate. 

\begin{enumerate}
    \setlength{\itemsep}{3pt}
    \setlength{\parskip}{0pt}
    \item Define a circular mask with radius $r$ in the activation grid to avoid edge effects during transformations.
    \item For the untransformed image $\mathbf{I}$, we define a sampling function $\mathcal{S}_{n,r}(\cdot)$ that selects $n$ activation coordinates within the circular mask, prioritizing those with the highest magnitude (L2 norm).
    \item Let $C_{\mathbf{I}} = {(x_1, y_1), ..., (x_n, y_n)}$ be the set of $n$ coordinates selected by $\mathcal{S}_{n,r}(\mathbf{a}^l(\mathbf{I}))$.
    \item For each transformed image $\mathbf{I}' \in \mathcal{G}(\mathbf{I})$, we apply the inverse transformation to the activation grid to maintain consistent sampling positions: $$\mathcal{T}_{\mathrm{g}^{-1}}(\mathbf{a}^l(\mathbf{I}'))$$ where $\mathcal{T}_{\mathrm{g}^{-1}}(\cdot)$ is the inverse of the transformation applied to create $\mathbf{I}'$. Then, for each sampled coordinate $C_{\mathbf{I}}$, we extract that coordinate from each set of transformed activations to create $x = [a^l(gI) ~:~ g\in G]$.
\end{enumerate}

Since we're focusing on the case of the dihedral group of order 32, our crosscoder will henceforth be referred to as a $\mathrm{D}_{32}$-crosscoder. It was trained on mixed3b, the final layer of InceptionV1’s “early vision” layers that have been well-characterised in prior work \cite{olah2020an}. It was trained with $10$ activations from each image for a single epoch with $\lambda = 3\times 10^{-7}$.

\begin{figure}[ht]
\vskip 0.2in
\begin{center}
\centerline{\includegraphics[width=\columnwidth]{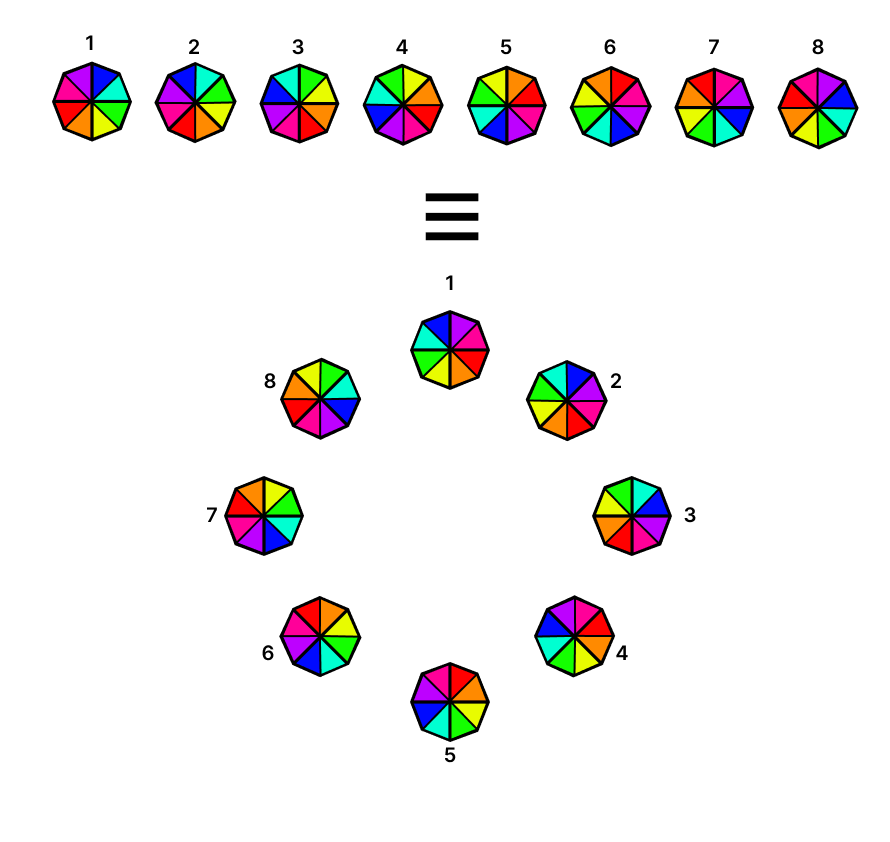}}
\caption{Using the rotated portion of $\mathrm{D}_8$ as an example, we can conceptualise the 1D vector as a circle.}
\label{figure:vector_conceptualisation}
\end{center}
\vskip -0.2in
\end{figure}

\subsection{Analysis}

\subsubsection{Distance Matrix Construction}\label{section:distance-matrix}

Two features, $i$ and $j$, are symmetrical if their dictionary vectors, $\mathbf{f}_i$ and $\mathbf{f}_j$, are related by some group action $\mathrm{g} \in \mathrm{G}$. 

To identify symmetrical features, we measure the maximum cosine similarity between two features under all actions $\mathrm{g} \in \mathrm{G}$:
$$
S_{ij} = \max_{g \in G_l} \cos\left( \mathbf{f}_i, \mathrm{g}(\mathbf{f}_j) \right),
$$
where $\cos(\cdot, \cdot)$ denotes the cosine similarity between two vectors, and $g(\mathbf{f}_j)$ represents the transformed feature after applying group action $g$ to $\mathbf{f}_j$.

\subsubsection{Group Operations on Dictionary Vectors}\label{section:group-operations}

We consider each dictionary vector $\mathbf{f}$ to be composed of $|\mathrm{G}|$ blocks of $n$ length, where each block corresponds to a specific transformation in $\mathrm{G}$. To perform operation $\mathrm{g}(\mathbf{f})$, we rearrange the “blocks” of $\mathbf{f}$, leaving the order of elements within each block unchanged.

The order of the activation blocks can be arbitrary, but for ease of explaining the operations, we specify the order as follows:
$$
\mathbf{f} = [e, \mathrm{r}, \mathrm{r}^2, \dots, \mathrm{r}^{2m}, ~~ \mathrm{s}, \mathrm{sr}, \mathrm{sr}^2, \dots, \mathrm{sr}^{2m}]
$$

Dictionary vectors can be conceptualised as two halves; the first corresponding to the rotations and the second corresponding to the reflections, and a visualisation of this structure can be found in Figure \ref{figure:vector_layout}.

\paragraph{Rotations.} To perform $\mathrm{r} \in \mathrm{D}_{2m}$, we perform a counter-clockwise circular shift within the rotation section and the reflection section. That is:
% $$
% \begin{align*}
% [e, \mathrm{r}, \mathrm{r}^2, \dots, \mathrm{r}^{2m}, ~~ \mathrm{s}, \mathrm{sr}, \mathrm{sr}^2, \dots, \mathrm{sr}^{2m}]\\
% ~~~\to 
% [\mathrm{r}, \mathrm{r}^2, \dots, \mathrm{r}^{2m}, e, ~~ \mathrm{sr}, \mathrm{sr}^2, \dots, \mathrm{sr}^{2m}, \mathrm{s}]
% \end{align*}
% $$

$$
\begin{gathered}
[e, \mathrm{r}, \mathrm{r}^2, \dots, \mathrm{r}^{2m}, ~~ \mathrm{s}, \mathrm{sr}, \mathrm{sr}^2, \dots, \mathrm{sr}^{2m}]\\
\to \\
[\mathrm{r}, \mathrm{r}^2, \dots, \mathrm{r}^{2m}, e, ~~ \mathrm{sr}, \mathrm{sr}^2, \dots, \mathrm{sr}^{2m}, \mathrm{s}]
\end{gathered}
$$

\paragraph{Reflections.} There are three goals to the reflection operation:
\begin{enumerate}
    \setlength{\itemsep}{3pt}
    \setlength{\parskip}{0pt}
    \item Perform a horizontal flip of $\mathrm{e}$.
    \item Maintain the order of the rotation and reflection block such that $\mathrm{s}_i$ represents the horizontal flip of $\mathrm{r}_i$, i.e., $\mathrm{s}_i = \mathrm{sr}_i$
    \item Preserve the relative order of rotations and reflections. Rotations occur counter-clockwise, whereas reflections “rotate” in a clockwise direction.
\end{enumerate}

\begin{figure*}[hbt!]
\begin{center}
\centerline{\includegraphics[width=\textwidth]{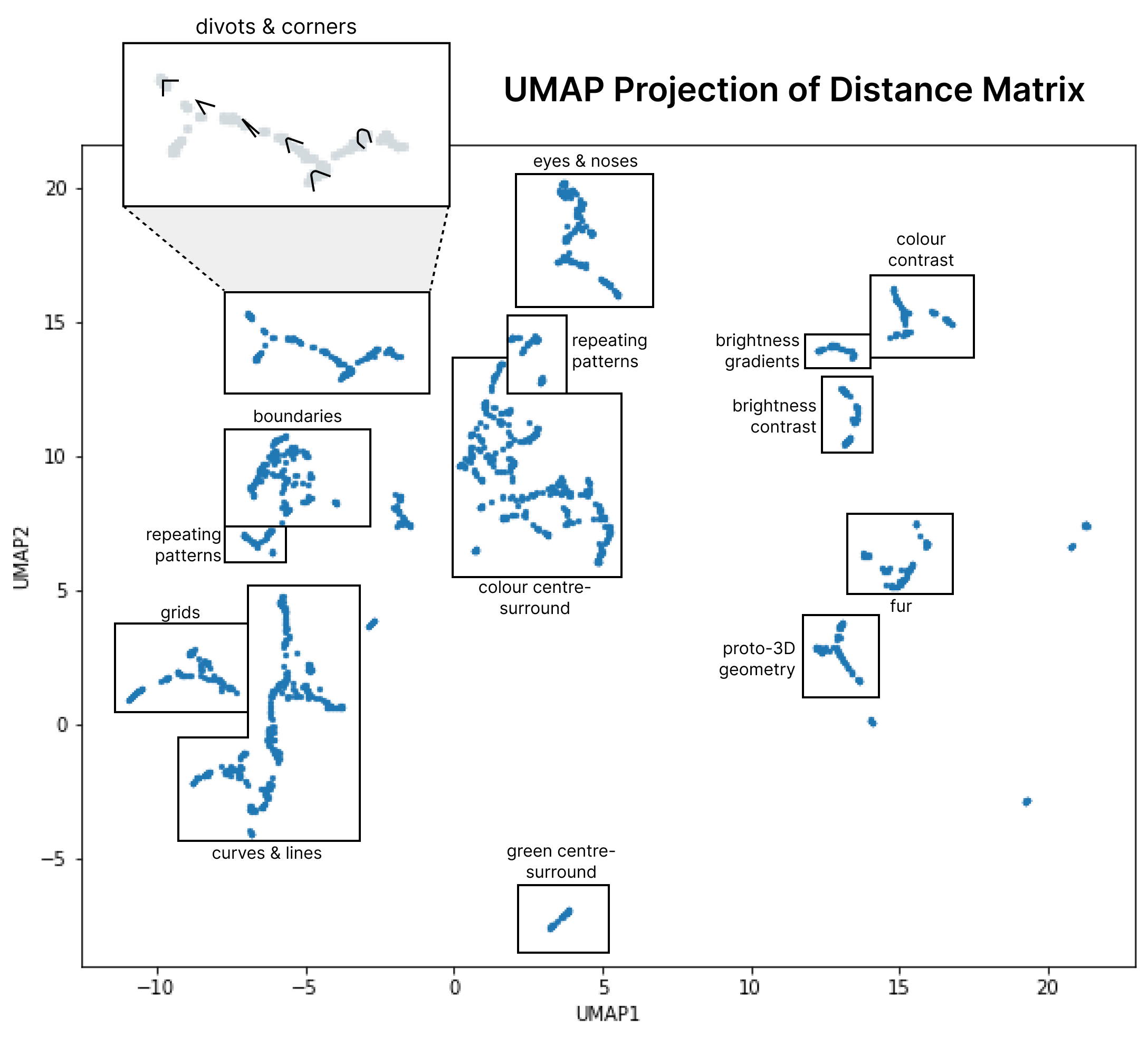}}
\caption{A 4D UMAP of group crosscoder features with the precomputed distance matrix from section \ref{section:distance-matrix} followed by a 2D UMAP with a cosine similarity metric. Distinct clusters of related features can be seen with structure emerging within some clusters, e.g., the "divots and corners" cluster.}
\label{figure:feature-umap}
\end{center}
\end{figure*}

To achieve this, we first swap the rotation and reflection sections:

$$
[\mathrm{s}, \mathrm{sr}, \mathrm{sr}^2, \dots, \mathrm{sr}^{2m}, ~~ e, \mathrm{r}, \mathrm{r}^2, \dots, \mathrm{r}^{2m}]
$$

Then, within the new reflection section, we reverse the order of all elements except this first, giving us:

$$
[\mathrm{s}, \mathrm{sr}, \mathrm{sr}^2, \dots, \mathrm{sr}^{2m}, ~~ e, \mathrm{r}^{2m}, \mathrm{r}^{2m-1}, \dots, r]
$$

This reversal is because, within both the rotation and reflection blocks, we can conceptualise them as a circle. That is, the final element in the block is only a single operation away from the first, as shown in Figure \ref{figure:vector_conceptualisation}. 

\begin{figure}[ht]
\vskip 0.2in
\begin{center}
\centerline{\includegraphics[width=\columnwidth]{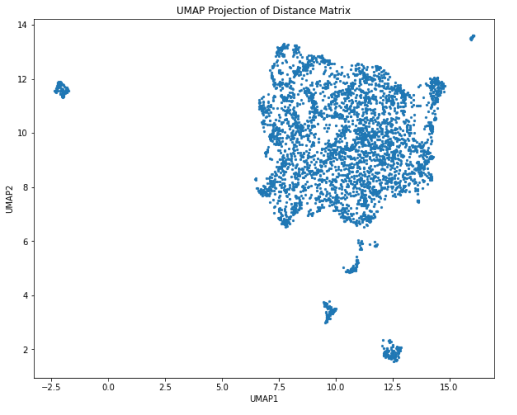}}
\caption{The dictionary vectors of a regular sparse autoencoder trained on the entirety of \texttt{mixed3b} following the methodology described in \citet{gorton2024missingcurvedetectorsinceptionv1}. Following the same UMAP procedure in \ref{figure:feature-umap} except using cosine similarity for the metric for both the 4D and the 2D UMAPs.}
\label{figure:sae-umap}
\end{center}
\vskip -0.2in
\end{figure}

\begin{figure*}[hbt!]
\begin{center}
\centerline{\includegraphics[width=\textwidth]{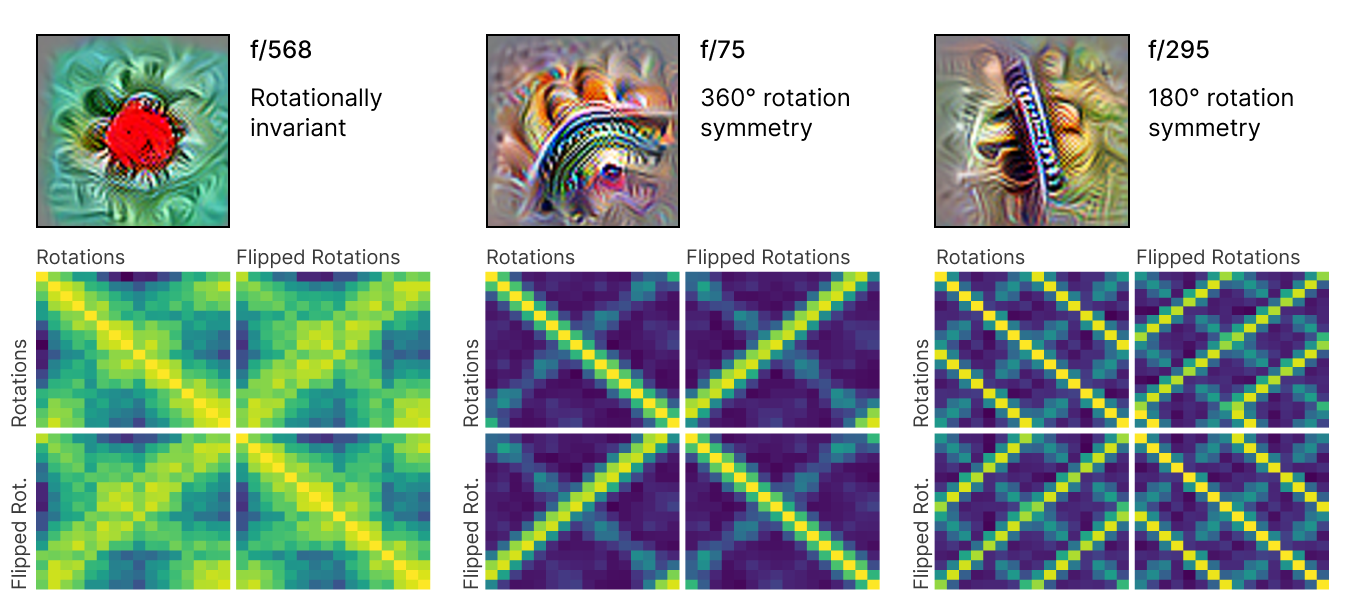}}
\caption{The cosine similarity of each block within three different features. For each feature, the top left and bottom right correspond to rotations, and the top right and bottom left correspond to reflections.}
\label{figure:cosine_sim_per_feature}
\end{center}
\end{figure*}

\subsubsection{Feature Visualisation}

We performed feature visualisation following \citet{olah2017feature} using our own fork of the Lucent library \cite{torch-lucent, gorton2024lucentfork}. Feature visualisation is performed using the encoder weight for a feature.\footnote{One of the advantages of training group crosscoders to reconstruct the transformations from only the untransformed image is that feature visualisation can proceed as it would with a normal sparse autoencoder. Feature visualisation becomes notably more computationally expensive if full sets of transformations are required as input.}

\section{Results}

\subsection{Feature Family Clustering}

Using the precomputed distance metric described in section \ref{section:distance-matrix}, we first performed a 4D UMAP and then a 2D UMAP using a cosine similarity metric (this best preserves the structure in high dimensional spaces). As shown in Figure \ref{figure:feature-umap}, group crosscoders produce interpretable clusters of features that largely correspond with this layer's previously hypothesised feature families.  Within each cluster, some additional structure emerges. For example, the “divots and corners” cluster appears as a spectrum of angularity and curvature. 

In contrast, if we consider a standard sparse autoencoder on InceptionV1’s \texttt{mixed3b}, we do find meaningful structure emerges within a family (e.g. within the curve detectors), but there isn’t as obvious a distinction between naturally occurring families of features. This can be seen in Figure \ref{figure:sae-umap}. Even compared to representations that are higher level and thus cleaner, such as those in InceptionV1's \texttt{mixed5b}, distinguishing between families of features is a lot fuzzier \cite{gorton2024interpretablefeaturesmixed5b}.

\subsection{Feature Symmetry Analysis}

One interesting property of group-crosscoders is that we can automatically analyse the symmetries and invariances of each feature with respect to our group. We take the dictionary vector, divide it into blocks corresponding to different group elements, and visualize the cosine similarities of the resulting vectors. Figure \ref{figure:transformer_block_cossim} shows the results of this across three features of decreasing curvature. The top left and bottom right represent rotations and then the top right and bottom left represent the reflections along the axes we rotate along.

Note how the curve detectors are only equivalent after a full 360\textdegree rotation, whereas the line feature \texttt{f/295} has two off diagonal lines corresponding to the fact that it's invariant to a 180\textdegree rotation.

\section{Conclusion}

We have introduced group crosscoders, a novel extension of crosscoders that enables the analysis of symmetrical features in neural networks. By training on transformations across a symmetry group rather than across layers, our approach provides several key advantages for mechanistic interpretability:

First, group crosscoders naturally discover and cluster related features into interpretable families, providing clearer and more objective separation between feature types compared to standard sparse autoencoders. This clustering emerges organically from the structure imposed by training on group transformations, suggesting that our method captures meaningful relationships between features that share symmetrical properties.

Second, our transform block analysis reveals how individual features respond to different transformations, offering quantitative insights into feature behaviour under various symmetries. This analysis demonstrates that features with different geometric properties (like curves versus lines) exhibit distinctly different patterns of similarity across transformations, matching our intuitive understanding of these features.

Our results on InceptionV1's mixed3b layer suggest that group crosscoders can help understand symmetry in neural networks. While we focused on dihedral groups and vision models, our approach can be made more general, both in using a different group (perhaps incorporating scaling or hue rotation) or applying to other modalities.

\section*{Acknowledgments}

I am grateful to Chris Olah for valuable discussions and feedback that helped refine the ideas presented in this work.

\bibliography{bibliography}
\bibliographystyle{icml2024}

%%%%%%%%%%%%%%%%%%%%%%%%%%%%%%%%%%%%%%%%%%%%%%%%%%%%%%%%%%%%%%%%%%%%%%%%%%%%%%%
%%%%%%%%%%%%%%%%%%%%%%%%%%%%%%%%%%%%%%%%%%%%%%%%%%%%%%%%%%%%%%%%%%%%%%%%%%%%%%%
% APPENDIX
%%%%%%%%%%%%%%%%%%%%%%%%%%%%%%%%%%%%%%%%%%%%%%%%%%%%%%%%%%%%%%%%%%%%%%%%%%%%%%%
%%%%%%%%%%%%%%%%%%%%%%%%%%%%%%%%%%%%%%%%%%%%%%%%%%%%%%%%%%%%%%%%%%%%%%%%%%%%%%%
\newpage
\appendix
\onecolumn
\section{Alternative Approaches}

Prior to the crosscoder work by \citet{lindsey2024crosscoders}, we had been working with a different approach, and although results will not be included, we believe it is valuable to describe that methodology.

InceptionV1 activations were gathered according to the methodology described in section \ref{section:dataset-training}.

We used sparse autoencoders we had trained previously on mixed3b. For a given feature, $i$, we construct a vector, $\mathbf{q}$, where $\mathbf{q}_{j, k}$ represents the feature’s activation, $\mathbf{f}_i$, on the $k$-th transform on the $j$-th image.

We can then utilise a very similar permutation strategy to section \ref{section:group-operations}, allowing us to perform group operations on a feature. The difference is that, given the concatenation across all the images, the permutation must be done per image. That is, the order of the images never changes; instead, each permutation is performed on each set of indices that correspond to an image.

The primary benefit to group crosscoders over this approach is that it front-loads the computationally expensive steps into training the model. The vectors for each feature get very long such that analysing them can be non-trivial. 

\end{document}